\documentclass[conference]{IEEEtran}
\IEEEoverridecommandlockouts
\usepackage{cite}
\usepackage{amsmath,amssymb,amsfonts}
\usepackage{algorithmic}
\usepackage{graphicx}
\usepackage{textcomp}
\usepackage{xcolor}
\usepackage{hyperref}
\usepackage{multirow}
\usepackage{todonotes}
\usepackage{subcaption}
\usepackage[dvipsnames]{xcolor}
\def\BibTeX{{\rm B\kern-.05em{\sc i\kern-.025em b}\kern-.08em
    T\kern-.1667em\lower.7ex\hbox{E}\kern-.125emX}}
\begin{document}

\title{General Semantic Knowledge Infusion for Spatio-Temporal Traffic Forecasting}

\author{
\IEEEauthorblockN{
Mattis thor Straten\IEEEauthorrefmark{1},
Yannick W\"olker\IEEEauthorrefmark{1}\IEEEauthorrefmark{2}, 
Steffen Strohm\IEEEauthorrefmark{1}, 
Prathvish Mithare\IEEEauthorrefmark{3},
Ralf Krestel\IEEEauthorrefmark{1}\IEEEauthorrefmark{3} and Matthias Renz\IEEEauthorrefmark{1}}

\IEEEauthorblockA{
\IEEEauthorrefmark{1}Department of Computer Science, Kiel University, Germany \\
\{mts, ywoe, sts, mr\}@cs.uni-kiel.de}
\IEEEauthorblockA{\IEEEauthorrefmark{2}GEOMAR Helmholtz Centre for Ocean Research Kiel, Germany \\}
\IEEEauthorblockA{\IEEEauthorrefmark{3}ZBW -- Leibniz Information Centre for Economics, Kiel, Germany\\
\{p.mithare, r.krestel\}@zbw.eu}
\thanks{© 2026 IEEE.  Personal use of this material is permitted.  Permission from IEEE must be obtained for all other uses, in any current or future media, including reprinting/republishing this material for advertising or promotional purposes, creating new collective works, for resale or redistribution to servers or lists, or reuse of any copyrighted component of this work in other works.}
}
\maketitle

\begin{abstract}
  Although Graph Neural Networks (GNNs) have made significant advances in spatio-temporal traffic forecasting, their performance is limited when they rely solely on sensor proximity or road-network topology. 
This paper presents a spatio-temporal prediction framework, developed to incorporate knowledge in various forms. 
This framework aims to improve sensor-level, contextual understanding of the environment. 
A general-purpose knowledge graph (e.g., Wikidata) is used to create semantic subgraphs around traffic sensors and generate knowledge graph embeddings that capture meaningful relationships, such as nearby points of interest, administrative hierarchies, and the functional roles of locations. 
These embeddings are then fused with conventional traffic sensor graphs to provide additional adjacency matrices informed by semantics. 
This allows GNNs to learn the semantic context beyond physical connectivity.
This study differs from previous research in two key ways.
Firstly, rather than proposing a novel GNN architecture, it demonstrates the general impact of external knowledge on prediction accuracy.
Secondly, experiments with well-established traffic forecasting approaches show that external knowledge provides additional information that street network data alone cannot convey. 
The results show that integrating data from general-purpose knowledge graphs and sensor networks through data fusion can enhance the prediction accuracy of traffic forecasting models, and offers a potential pathway toward improved interpretability.
\end{abstract}


\section{Introduction}
Modern urban environments are becoming increasingly equipped with dense sensor infrastructures that monitor mobility, infrastructure and environmental conditions. 
However, the data generated by these systems is often distributed across heterogeneous sources.
Often, predictive analytics for urban dynamics, such as traffic prediction, solely focus on one of these sources.
Knowledge infusion, which involves integrating structured contextual knowledge into data-driven analytics, has therefore emerged as a promising approach~\cite{zheng_fusing_2026, zheng_methodologies_2015}.
Crossing the borders between these heterogeneous sources is key to improving the performance and interpretability of spatio-temporal predictions~\cite{he_geolocation_2025}. 
Traffic prediction in particular benefits from the semantic context, as vehicle flows are not solely determined by spatial proximity, but also by the functional characteristics of the surrounding urban environment~\cite{GMAN-AAAI2020, wu_graph_2019, shao_decoupled_2022}.

While the road network of a city is a two-dimensional, spatially embedded structure, the latent factors influencing traffic are inherently multi-perspective (e.g., land use patterns, nearby points of interest, administrative hierarchies or street semantics). 
Such multifaceted context can be formally represented through knowledge graphs (KGs)~\cite{hogan_kg_2021}, which provide structured relationships among places, entities, and concepts. 
However, constructing domain-specific KGs tailored to individual cities requires significant manual effort and adaptation, which limits scalability and results in poor generalizability across different urban contexts~\cite{zhang_context-aware_2025}.
Leveraging large-scale, general-purpose KGs (e.g., Wikidata) offers a practical, transferable alternative that enables knowledge infusion without bespoke crafting for each region or restriction to a single use case.

In Graph Neural Network–based traffic prediction approaches, the adjacency matrix serves as the interface for infusing external knowledge into the prediction mechanism~\cite{zhang_context-aware_2025, zhu_kst-gcn_2022}.
Current state-of-the-art techniques commonly define this adjacency matrix using spatial properties, such as travel time or network distance.
This work investigates the impact of semantically derived adjacency matrices constructed from general-purpose KGs on existing prediction approaches. 
The advantage of a curated general-purpose KG is its explainability and steadiness compared to LLM-based systems that could create such semantical connections~\cite{hajisafi_learning_2023}. 
However, the goal of this study is to isolate the effect of the adjacency matrix on the traffic prediction, which can be utilized in the future to induce knowledge gathered by AI systems.
Specifically, this work contributes the following:

\begin{itemize}
    \item A general-purpose pipeline\footnote{\url{https://github.com/cau-kiel-ai/Semantic-Knowledge-Infusion-ST-Prediction}} for knowledge infusion, not relying on hand-crafted semantic features or curated KGs.
    
    \item A systematic study of the semantic relationships derived from KGs in established traffic prediction approaches.

    \item A controlled analysis that allows attributing performance gains to added semantic context.
\end{itemize}







\begin{figure*}
    \centering
    \includegraphics[width=0.8\linewidth]{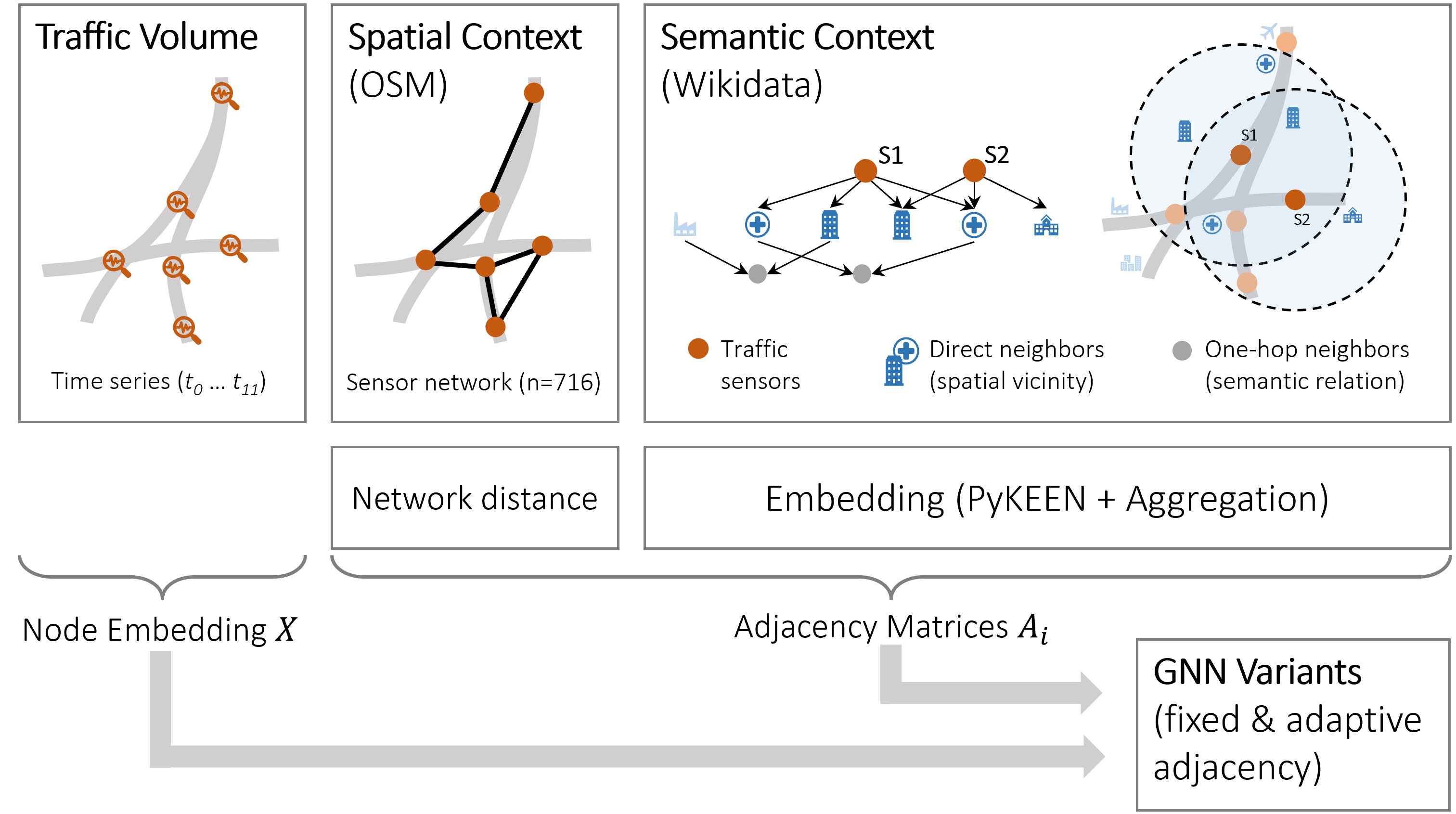}
    \caption{Method overview: The considered traffic datasets are represented by past traffic volume measurements at specific locations within the traffic network. 
    These measurements serve as input for spatio-temporal GNNs, which are used to predict future timesteps. 
    Based on specific locations (orange), two contexts or perspectives of sensor connectivity are created. 
    The \textit{Spatial Context} is commonly used and represents network distance based on OpenStreetMap estimates.
    The \textit{Semantic Context} connects the locations of the sensors with the Wikidata KG to extract the surrounding semantics, which are embedded using the PyKEEN library to compute the semantic similarity between any pair of sensors. 
    Both contexts provide adjacency matrices that are used in spatio-temporal GNNs to determine the flow of node embeddings containing traffic volume.}
    \label{fig:workflow-overview}
\end{figure*}

\section{Related Work}

\subsection{Traffic Prediction with Graph Neural Networks}
The traffic prediction problem, when not considering knowledge infusion, is typically described as a multivariate time series forecasting task~\cite{peng_graphrag_2024}. 
Given a set of sensors distributed throughout a traffic network (as depicted exemplarily in Figure \ref{fig:map}) and their past measurements, the objective is to predict the future traffic values for all sensors.
These traffic networks, which contain mostly highway data, are not regularly structured, making the usage of graph structures a good choice to handle the spatial distribution\cite{han_bigst_2024, yu_spatio-temporal_2018, li2018dcrnn_traffic, wu_graph_2019, shao_decoupled_2022, xu_spatial-temporal_2020}.
Graph neural networks can model directional relationships between irregularly structured sensors based on graph connectivity~\cite{atluri_spatio-temporal_2019}.
For this reason, they have achieved significant success in traffic prediction.

In general, each traffic sensor is modeled as a graph node in GNNs such that the \textit{adjacency matrix} determines connectivity and information exchange.
Early graph-based approaches rely on fixed adjacency matrices that encode the distance of the sensors in the road network with a threshold to sparsify the connections~\cite{yu_spatio-temporal_2018, li2018dcrnn_traffic}.
However, the field has developed in the direction of adaptive adjacency matrices, which are subject to learned weights during the training process. 
GWaveNet~\cite{wu_graph_2019} found that the adaptive adjacency matrix alone performs similarly to the spatial distance-based adjacency matrix, justifying a paradigm shift in the recent literature towards the adaptive adjacency matrix.
However, approaches using the adaptive adjacency matrix still use the spatial distance-based  adjacency matrix as a starting point~\cite{shao_decoupled_2022} or as an addition~\cite{wu_graph_2019, zhang_context-aware_2025}.
In general, the adjacency matrix is somewhat interpretable.
Therefore, it is a useful connector for further knowledge about the context of the predictive task.

\subsection{Knowledge Graph Embeddings}
The overall goal is to integrate symbolic knowledge representations into the predictive spatio-temporal task. 
Breit et al.~\cite{breit_MLSW_2023} identified KGs as central resources for enriching downstream prediction tasks through their structured semantics. 
To make KGs usable for this task, knowledge graph embeddings (KGE) provide a numerical representation of the entities and relations included in the KG.
Current surveys emphasize that incorporating multiple perspectives and ensuring interpretability and robustness are challenges for many KGE models~\cite{biswas_KGE_2023}.
Several libraries and frameworks have been developed to support reproducible and scalable KGE research~\cite{ali_pykeen_2020, broscheit_libkge_2020, zheng_dlkge_2020}.

Beyond classic embedding pipelines, graph-structured information is becoming an integral part of modern retrieval-augmented generation (RAG)~\cite{GraphRAG} systems.
Recent work on GraphRAG~\cite{grag, mavromatis-karypis-2025-gnn} demonstrates how graph-based retrieval can support downstream generation tasks by leveraging relational structure instead of relying solely on unstructured text retrieval.
This line of research underscores the growing relevance of KGs and their embeddings for enhancing reasoning and contextualization in large language model–based systems \cite{peng_graphrag_2024}.

\subsection{Knowledge Graphs for Traffic Prediction}
Predictive analytics in spatio-temporal domains based on neural network techniques are, by construction, not easily explainable~\cite{pmlr-v151-agarwal22b}. 
Hence, the infusion of machine-readable and human-understandable knowledge is also a major goal of previous studies.
It has shown that external knowledge sources can also help with data sparsity~\cite{wolker_suster_2023} and the reduction of computational resources~\cite{wolker_small_2025}.
Many previous approaches with this goal have two common features: a hand-designed KG~\cite{gong_empowering_2023} and a novel prediction architecture. 
For example, Zhang et al.~\cite{zhang_context-aware_2025} designed a KG with a set of dedicated node types (e.g., points of interest (POIs) and streets) and task-specific relation types. 
Due to the lack of hand-crafted datasets combining traffic observations and further spatial information, the KR-STGNN~\cite{kr_stgcn} and KST-GCN~\cite{zhu_kst-gcn_2022} frameworks are evaluated on a single dataset with data from Shenzhen, CN.
The dataset includes weather data for one month and nine types of POIs, which were used to create the KG manually.
This limits the application of the approaches to different cities.

Wang et al.~\cite{kfgnn} introduced knowledge infusion to established traffic prediction baselines using two different traffic datasets. 
They did this by adding semantic adjacency matrices, such as traffic structure, traffic pattern, and regional functionality matrices.
However, their study did not investigate how the extraction process could be generalized or its influence on prediction accuracy.

\section{Method}
This work enriches traffic forecasting by incorporating semantic context from a general-purpose KG in the region of interest. 
This is accomplished by modifying the adjacency matrices of existing approaches (see Figure \ref{fig:workflow-overview}). 
The presented approach utilizes geo-referenced entities in the KG located spatially close to traffic sensors to create an approximate semantic context of the sensors using KGE.
Thus, the pairwise similarity between the neighborhoods of two sensors can be leveraged as a semantic adjacency matrix.

\subsection{Spatially Connected Knowledge Graph}
The spatial information of the traffic sensors is used to construct a spatially bounded subgraph of the selected KG by identifying KG entities within a predefined distance threshold -- representing spatially relevant POIs (see \textit{Direct neighbors} in Figure \ref{fig:workflow-overview}).
This distance-based filtering defines the contextual neighborhood of each sensor within the KG and ensures that the resulting subgraph only contains entities that are spatially relevant to the sensors.
Two different subgraphs are created based on the spatially relevant POIs (see Figure \ref{fig:workflow-overview}): 
\begin{enumerate}
    \item[]\textbf{Direct Neighborhood}. Contains POIs and their respective types (e.g., restaurant, shopping mall), as well as all KG properties (edges) that connect POIs and types.
    These properties contain relations between POIs, the allocation of types to POIs and taxonomic super-/subtype relationships, thereby creating an induced subgraph of the selected KG.
    \item[]\textbf{One-Hop Neighborhood}. The \textit{Direct Neighborhood} subgraph with the one-hop neighbors of the POIs and the types of all entities in the subgraph added. 
    All properties that establish interconnections between any two entities contained within the selected KG are also included to create an induced subgraph of the selected KG.
\end{enumerate}

The gathered KGs contain different semantic information about POIs that are potentially relevant to the traffic sensors.
This work uses the \textit{ComplEx} model~\cite{Trouillon_ComplEx_2016} for KGE, which is well suited for capturing the asymmetric relations that frequently occur in KGs.
Using this KGE method on the extracted subgraph creates a numerical, complex-valued representation of each KG entity.
The average aggregation of the embeddings of all POIs spatially close to a traffic sensor represents the sensor's semantic embedding vector, allowing to incorporate semantic KG information into the traffic prediction.
Using an average aggregation yields a fixed-size, order-invariant representation that is independent of the absolute number of nearby POIs and robust to heterogeneous POI types, while more expressive aggregation schemes (e.g., weighted or attention-based pooling) are left for future work.


\subsection{Spatially Infused Traffic Prediction}
The adjacency matrix serves as an interface between the real world and traffic prediction models.
Accordingly, the embedding vectors of traffic sensors are used to derive a semantically informed adjacency matrix (see Figure~\ref{fig:workflow-overview}).
Semantic similarity between sensor entities is quantified using a cosine-inspired similarity measure derived from their complex-valued embeddings created by the ComplEx model.
Following the ComplEx formulation~\cite{Trouillon_ComplEx_2016}, similarity is based on the Hermitian inner product between two embeddings, where the real part of this product is retained to obtain a real-valued score. 
The resulting similarity measure is normalized by the vector norms, yielding a cosine-like similarity for complex-valued embeddings.
The symmetric, normalized adjacency matrix $A$ with entries $a_{ij}$ is computed between two complex-valued sensor embeddings $\mathbf{e}_i, \mathbf{e}_j \in \mathbb{C}^d$ of sensors $i$ and $j$ as follows:

\begin{equation}
    a_{ij} = \frac{\Re(\sum_k \overline{\mathbf{e_i}_k} \mathbf{e_j}_k)}{||\mathbf{e_i}||~||\mathbf{e_j}||}
\end{equation}

However, the semantic adjacency matrix is not independent of the spatial adjacency matrix. 
Sensors in close spatial proximity are expected to exhibit highly similar semantic footprints and share many neighboring entities.
Therefore, the semantic adjacency matrix shows spatial autocorrelation patterns in addition to long-distance relationships that are solely semantically motivated (e.g., business districts and vacation resorts).

The included experiments require the selected approaches to handle multiple adjacency matrices, which can be interpreted as a set of filters applied to the input traffic volume.
The application of multiple adjacency matrices across many established approaches can be summarized as follows:
Given a set of adjacency matrices $\mathcal{A}$ and a graph convolution function $GC$, that takes an input matrix $X$ (e.g., traffic volume or hidden representations), an adjacency matrix $A_i \in \mathcal{A}$ and learnable weights $W_i$ as input, the resulting convoluted features $H$ are defined as:

\begin{equation}
    H = W \times \Large{\mathop{\|}\limits_{\small A_i \in \mathcal{A}}}\normalsize GC(X, A_i, W_i) 
    \label{eq:concat_gcn}
\end{equation}

where $\mathop{\|}$ is the concatenation operator and $W$ is a learnable weight matrix transforming the concatenated embeddings for each graph node in the target dimension for the architecture. 
This allows the cardinality of $\mathcal{A}$ to change, giving the architecture a new opportunity to use additional semantic information as long as the cardinality is known prior to the training.
 \label{chap:methods}
\section{Results}





This work demonstrates the effectiveness of integrating semantic information into preexisting traffic prediction models, and compares the results with those of unmodified approaches.

\subsection{Experiment Data \& Setup}
\begin{figure}
    \centering
    \includegraphics[width=0.8\linewidth]{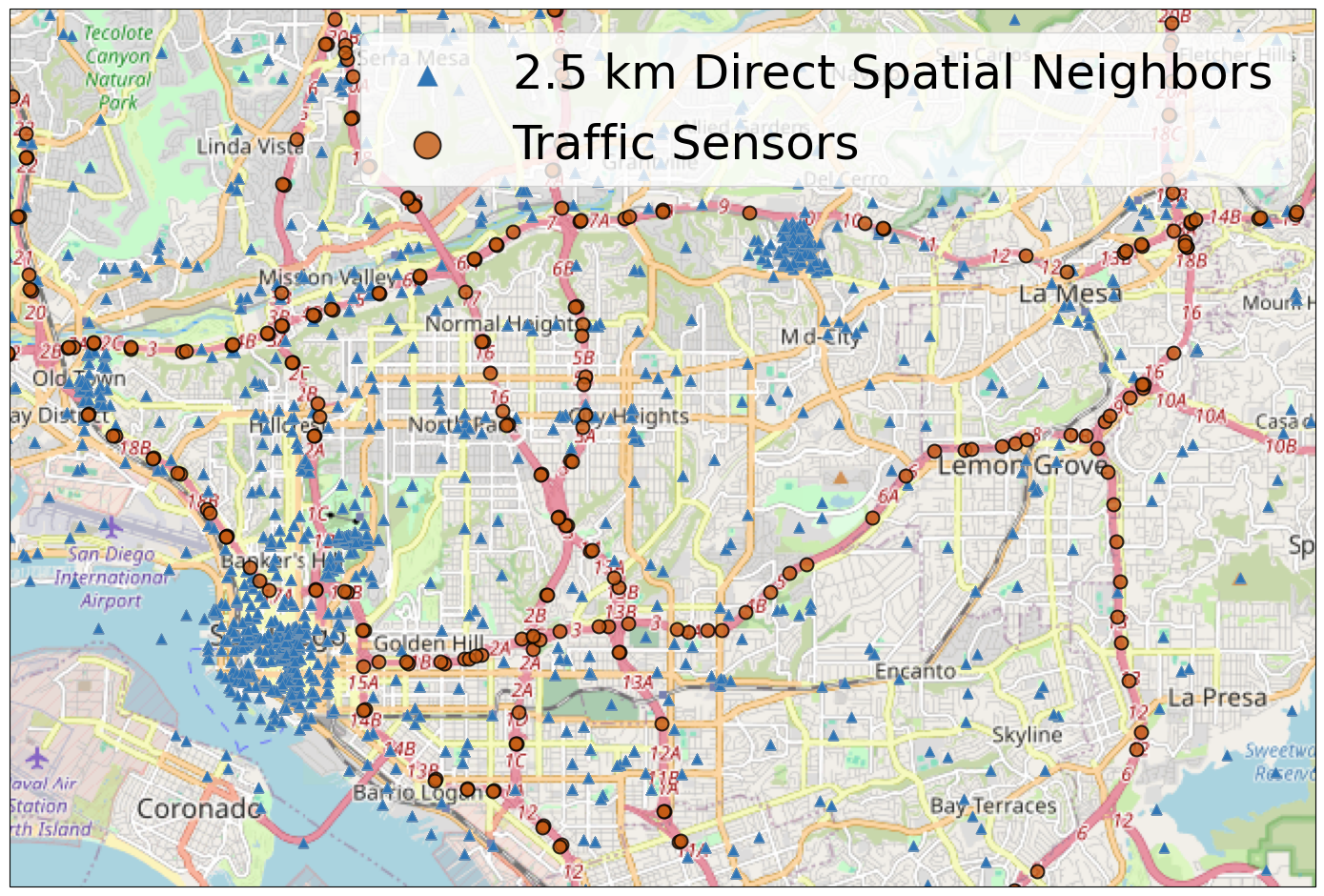}
    \caption{OpenStreetMap excerpt of traffic sensors from San Diego and Wikidata POIs within a 2.5 km radius.}
    \label{fig:map}
\end{figure}

For the evaluation, a highway traffic dataset from San Diego, USA, which is part of the LargeST benchmark \cite{liu_largest_2023}, is employed. 
The dataset comprises 716 traffic sensors (circular icons in Figure~\ref{fig:map}) and more than 35\,000 temporal measurements.
The spatial adjacency matrix is constructed using a Gaussian kernel applied to network distances, with edges retained only if the resulting kernel value exceeds 0.01.
This results in approx. 3\% of the adjacency matrix being filled. 
Further details are provided in the original LargeST publication.

\paragraph{Embedded Wikidata Subgraphs}
The publicly available, community-maintained Wikidata KG is queried spatially to gather contextual information around the sensors due to its large-scale and cross-domain coverage \cite {vrandecic_wikidata_2014}.
However, as a collaboratively curated resource, Wikidata exhibits heterogeneous coverage and varying data quality across global regions, which may make other general-purpose knowledge graphs more useful for some datasets.
In this work, geo-referenced Wikidata \textit{entities} that are spatially close to any traffic sensor are retrieved.
Experimental thresholds of 600 m and 2.5 km are chosen to define spatial closeness.
The 600 m threshold captures the immediate urban surroundings of a San Diego traffic sensor -- typically the scale of nearby intersections or road infrastructure that directly influences local traffic flow.
In contrast, the 2.5 km threshold reflects a broader mobility context, as shown in Figure \ref{fig:map}, including major attractors or traffic propagation effects operating at the neighborhood scale. 
Using both thresholds enables the analysis to capture spatial effects at different scales, ranging from direct, local influences to more diffuse, neighborhood-level interactions that are potentially relevant for traffic dynamics.

Based on the spatially selected Wikidata entities, the two subgraphs described in chapter \ref{chap:methods} are created.
The \textit{Direct Neighborhood} subgraph finds the associated types, e.g., \textit{university} (Q3918), using \textit{instance of}(P31). 
Two types are interconnected by \textit{subclass of} (P279) and connections between Wikidata entities are established using any \textit{direct properties} in the KG connecting them.
The \textit{One-Hop Neighborhood} subgraph adds one-hop neighbors and their types, considering direct Wikidata properties that start or end at a direct neighbor. 
Two types are interconnected using the \textit{subclass of}-property, and two other entities are connected using any direct Wikidata property that relates them.
The sizes of the created subgraphs are shown in Table \ref{tab:subgraph_sizes}.

\begin{table}[ht]
\centering
\caption{Size of extracted subgraphs for the distance thresholds of 600 m and 2.5 km, considering the Direct Neighbors (D) and One-Hop Neighbors (1H) subgraphs.}
\label{tab:subgraph_sizes}
\begin{tabular}{c r r r r}
\hline
& \multicolumn{2}{|c}{600 m} & \multicolumn{2}{|c}{2.5 km} \\
Subgraph & \multicolumn{1}{|r}{Nodes} & Edges (Triples) & \multicolumn{1}{|r}{Nodes} & Edges (Triples) \\
\hline
D &   797  &   1\,827 &  2\,770 &   8\,062 \\
1H & 2\,920 &  12\,523 & 29\,655 & 144\,141 \\
\hline
\end{tabular}
\end{table}

This study uses PyKEEN~\cite{ali_pykeen_2020} to embed the Wikidata subgraphs, due to its composable pipeline for interaction models, loss functions, training strategies, and automatic hyperparameter optimization. 
Based on preliminary experiments outside the scope of this paper, a 64-dimensional embedding space is selected to adequately represent subgraphs containing up to 144\,141 triples (see Table~\ref{tab:subgraph_sizes}).
This dimensionality balances representational capacity for heterogeneous relations with computational efficiency, and is fixed across all experiments to ensure comparability and avoid additional degrees of freedom.
All available triples are used for training, as embeddings are learned in a self-supervised manner with the goal of obtaining coherent semantic representations rather than optimizing predictive performance on held-out data.
As described in chapter \ref{chap:methods}, the embeddings of all POIs within 600 m or 2.5 km of a sensor are aggregated to produce sensor-level vectors.

\paragraph{Traffic Prediction Baselines}
For traffic prediction baselines, approaches capable of handling multiple adjacency matrices, as described in Eq.~\ref{eq:concat_gcn}, are employed.
The following baseline methods, using the implementations provided by the LargeST benchmark~\cite{liu_largest_2023}, are included:
\begin{itemize}
    \item STGCN~\cite{yu_spatio-temporal_2018} is a convolution-based approach in both time and space. It uses fixed adjacency matrices of the road network.
    \item DCRNN~\cite{li2018dcrnn_traffic} works with fixed adjacency matrices in an encoder-decoder architecture, where graph convolutions are part of the temporal gating mechanism.
    \item GWaveNet~\cite{wu_graph_2019} utilizes a learnable node embedding to create a learnable adjacency matrix, named \textit{GWaveNet a.}, and a variant with solely fixed adjacency matrices, as both were part of the original paper.
    \item D2STGNN~\cite{shao_decoupled_2022} decouples the inherent signal from the diffusion signal. This approach uses, similar to GWaveNet, a node embedding for the prediction, which can be extended by fixed adjacency matrices. 
    \item STTN~\cite{xu_spatial-temporal_2020} is a Transformer architecture using spatial attention between the sensors, comparable to the node embeddings. Here, the fixed adjacency matrix can be seen as a constant attention score distribution.
\end{itemize}

\paragraph{Hardware}
Throughout the experiments, evaluations are conducted on a high-performance computing system: 8-core Intel Xeon Gold 6226R CPU, 16 GB of memory, and an NVIDIA V100 GPU (32 GB).

\begin{table*}[]
\footnotesize
    \centering
    \caption{Mean Absolute Error of traffic prediction skill when replacing the spatial adjacency matrix with the semantic adjacency matrices and a random adjacency matrix. The variance is reported for the original runs.}
    \begin{tabular}{l|l|r|r||r|r|r|r}
         Adjacency & Baselines & Original & Random & 600m D & 600m 1H & 2.5km D & 2.5 1H\\
         \hline

\multirow{3}{*}{Fixed} & STGCN & 20.91{\scriptsize $\pm$0.98} & 20.13 (-3.7\%) & 19.97 (-4.5\%) & 19.83 (-5.1\%) & 19.65 (-6.0\%) & \textbf{19.61 (-6.2\%)} \\
& DCRNN & 30.84{\scriptsize $\pm$46.31} & \textbf{27.12 (-12.1\%)} & 33.28 (+7.9\%) & 34.00 (+10.3\%) & 28.67 (-7.0\%) & 38.52 (+24.9\%) \\
& GWaveNet & \textbf{19.15}{\scriptsize $\pm$0.19} & 19.45 (+1.6\%) & 25.16 (+31.4\%) & 24.24 (+26.6\%) & 20.18 (+5.4\%) & 21.98 (+14.8\%) \\
\hline
\multirow{2}{*}{Adaptive} & GWaveNet a. & \textbf{18.32}{\scriptsize $\pm$0.02} & 18.72 (+2.1\%) & 18.71 (+2.1\%) & 18.69 (+2.0\%) & 18.71 (+2.1\%) & 18.63 (+1.7\%) \\
& D2STGNN& 20.82{\scriptsize $\pm$0.00} & 18.07 (-13.2\%) & \textbf{17.64 (-15.2\%)} & 18.06 (-13.2\%) & 18.01 (-13.5\%) & 17.86 (-14.2\%) \\
& STTN & 19.27{\scriptsize $\pm$0.23} & 19.16 (-0.6\%) & 18.97 (-1.6\%) & 18.92 (-1.8\%) & \textbf{18.76 (-2.7\%)} & 19.03 (-1.2\%) \\

    \end{tabular}
    \label{tab:replaceResults}
\end{table*}

\begin{table*}[]
\footnotesize
    \centering
    \caption{Mean Absolute Error of traffic prediction skill when considering both the semantic and spatial adjacency matrices.}
    \begin{tabular}{l|l|r||r|r|r|r}
         Adjacency & Baselines & Original & 600m D & 600m 1H & 2.5km D & 2.5 1H\\
         \hline

\multirow{3}{*}{Fixed} & STGCN & 20.91 & 19.66 (-6.0\%) & \textbf{19.30 (-7.7\%)} & 19.67 (-5.9\%) & 19.62 (-6.2\%) \\
& DCRNN & 30.84 & 66.26 (+114.8\%) & \textbf{25.93 (-15.9\%)} & 54.73 (+77.5\%) & 27.23 (-11.7\%) \\
& GWaveNet & 19.15 & \textbf{18.75 (-2.1\%)} & 18.76 (-2.0\%) & 18.98 (-0.9\%) & 18.90 (-1.3\%) \\
\hline
\multirow{2}{*}{Adaptive} & GWaveNet a. & 18.32 & \textbf{18.17 (-0.9\%)} & 18.23 (-0.5\%) & 18.26 (-0.4\%) & 18.21 (-0.6\%) \\
& D2STGNN & 20.82 & 18.61 (-10.6\%) & \textbf{18.24 (-12.4\%)} & 20.44 (-1.8\%) & 19.87 (-4.5\%) \\
& STTN & 19.27 & 18.72 (-2.8\%) & 18.53 (-3.8\%) & \textbf{18.30 (-5.0\%)} & 18.53 (-3.9\%) \\
    \end{tabular}
    \label{tab:addResults}
\end{table*}

\subsection{Influence of Adjacency Matrix}
First, the influence of semantic adjacency matrices derived from the two different subgraphs, with each having two different spatial radii (see Table~\ref{tab:subgraph_sizes}), on the selected baseline models is investigated.
Besides all being able to handle multiple adjacency matrices, the chosen baselines differ in design choices depending on whether they use an adaptive adjacency matrix driven by the backpropagation learning process.
Since each baseline utilizes at least a fixed adjacency matrix with spatial context, two injection strategies are tested: replacing the spatial context (network distance adjacency matrix) with the semantic context (introduces semantic adjacency matrix), and additionally injecting the semantic context alongside the spatial adjacency matrix.
Tables \ref{tab:replaceResults} and \ref{tab:addResults} show the results as an average of 11 runs.
The \textit{Original} column reports the traffic prediction skill of the baselines for unchanged connectivity.
The results obtained for the unmodified baseline models largely align with the reported performance of the LargeST benchmark framework~\cite{liu_largest_2023}, which serves as the basis for this work.
The performance of incorporating the different semantic adjacency matrices is compared to that of the unmodified baseline model by showing the percentage change in mean absolute error.

\subsubsection{Only Semantic Context}
The results in Table~\ref{tab:replaceResults} are obtained by replacing the physical adjacency matrix, which describes distance-based relationships between traffic sensors, with the different semantic adjacency matrices.
Additionally, the physical adjacency matrix is replaced with a random adjacency matrix that is the same across all baselines and has the same density as the physical adjacency matrix, but connects random traffic sensors.

For the adaptive baselines D2STGNN and STTN, as well as the fixed baseline STGCN, all replaced semantic adjacency matrices improve the prediction skill compared to the baseline (network distance adjacency matrix) and the randomly chosen adjacency matrix.
However, no consistent improvement can be observed for GWaveNet and DCRNN when using the semantic adjacency matrix as a replacement.
This suggests that the usefulness of the semantic adjacency matrix depends on the approach's architecture.
Similar observations can be made when the physical adjacency matrix is replaced with a random adjacency matrix. 
In this case, most baselines also improve with a random matrix, complicating the attribution of gains to semantic content.

\subsubsection{Semantic and Spatial Context}
When the semantic adjacency matrix is included as an additional input, the results in Table~\ref{tab:addResults} demonstrate performance improvements for most traffic prediction baselines, with notable exceptions for DCRNN.

Even the GWaveNet approach, which did not benefit from the replaced semantic adjacency matrix, improved its prediction skills for both its fixed and adaptive versions across all four added semantic adjacency matrices.
Although most baselines diverge significantly from their original performance, the GWaveNet baseline variants benefit less from the semantic context.
Neither of the two variants improves when the adjacency matrix is replaced by the semantic one, and the improvement is limited when the semantic adjacency matrix is treated as an additional matrix: -2.1\% for the fixed variant and -0.9\% for the adaptive variant.
This highlights that the GWaveNet approach, in both its fixed and adaptive matrix variants, only has a limited improvement potential.
Conversely, the significant improvement in prediction skill of the other four baselines suggests that the choice of adjacency matrix is crucial.
Furthermore, these results suggest that the original adjacency matrix, which considers only network distance, provides limited information that could be expanded upon to improve performance. 
The semantic and original matrices can be understood as complementary. 
While the spatial context only connects spatially close traffic sensors, semantic adjacency can connect similar functional regions, such as residential areas or office building blocks.

\begin{figure}
    \centering

    \begin{subfigure}{\linewidth}
        \centering
        \includegraphics[width=0.9\linewidth]{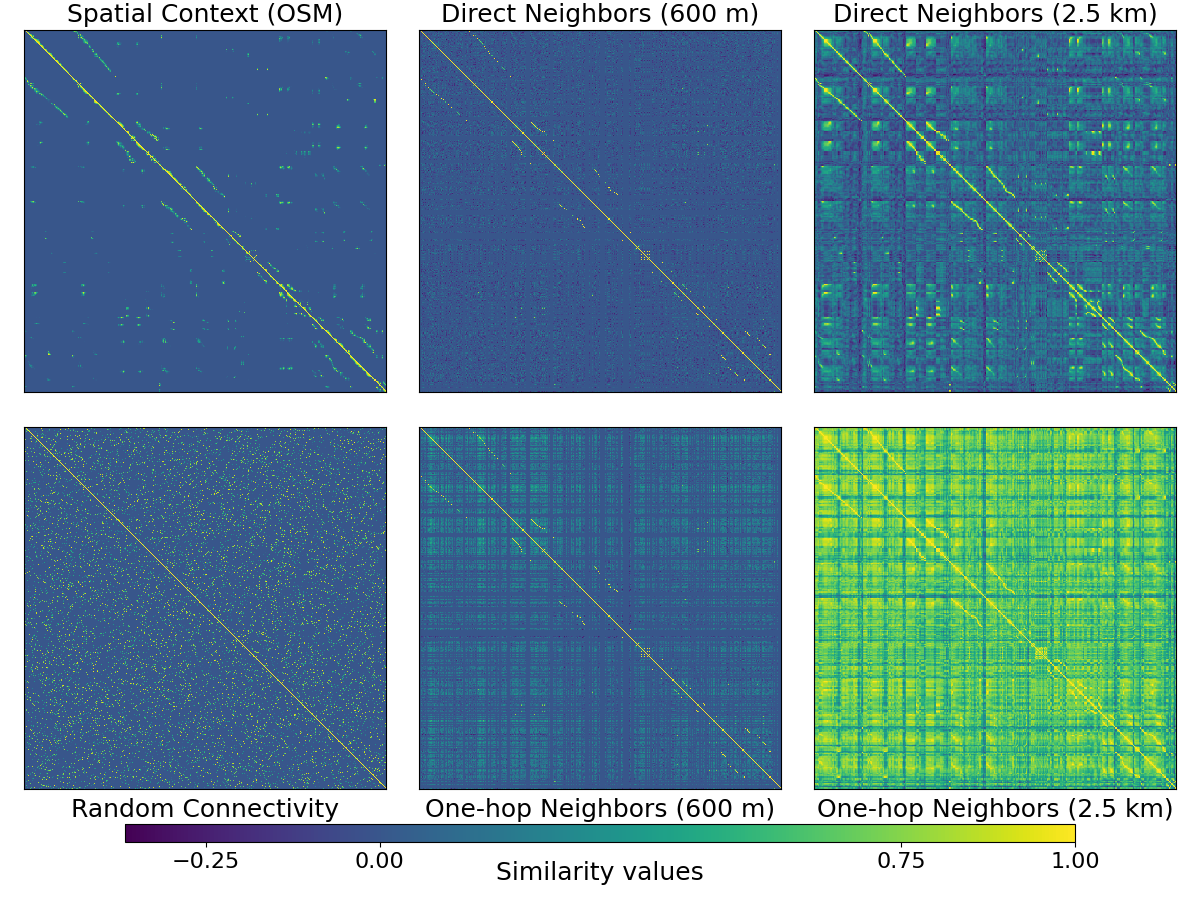}
        \caption{The similarity values of different adjacency matrices.}
        \label{fig:adjacency_matrices}
    \end{subfigure}


    \begin{subfigure}{\linewidth}
        \centering
        \includegraphics[width=0.9\linewidth]{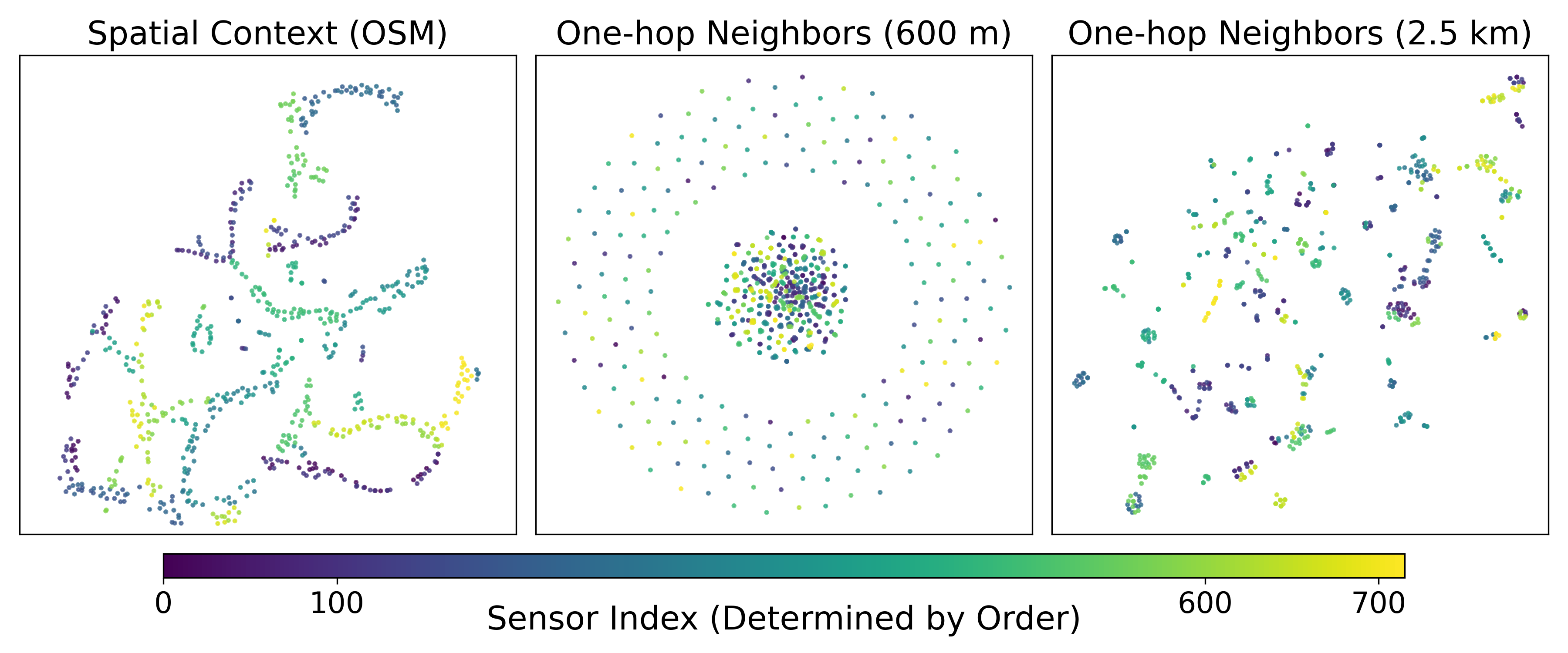}
        \caption{t-SNE plots of different adjacency matrices.}
        \label{fig:tsne}
    \end{subfigure}

    \caption{Structural aspects of different adjacency matrices 
    , incorporating spatial and semantic context.}
    \label{fig:matrix_plots}
\end{figure}

\subsubsection{Structure of Adjacency Matrices}
Figure \ref{fig:adjacency_matrices} visualizes the different adjacency matrices used throughout this work. 
The large performance divergence of the approaches based on the choice of the adjacency matrix indicates major differences in semantic and spatial contexts.
The spatial context (OSM) connects traffic sensors based on network distance, resulting in predominantly local connections that appear along the main and secondary diagonals due to the sensor ordering.
Similar patterns are observed in the semantic adjacency matrices, particularly for the larger radius of 2.5 km, where increased semantic overlap between neighboring traffic sensors potentially includes parts of the spatial context. 
Additionally, the semantic context introduces far more connections in areas of the adjacency matrix that are not covered by the spatial context.
This suggests that traffic sensors farther apart than the spatial threshold used to create the spatial adjacency matrix are connected.
The two-dimensional t-SNE~\cite{tsne} visualization (see Figure~\ref{fig:tsne}) of the adjacency matrices for the spatial and semantic one-hop neighbors for 600 m and 2.5 km shows the aforementioned pattern.
Considering the spatial matrix, sensors clustered together are colored similarly, indicating similar positions in the adjacency matrix, and thus showing the diagonal structure. 
The two-class separation of the 600 m semantic adjacency matrix is formed by traffic sensors that either have or do not have any spatial neighbors in the KG. 
Interestingly, the clusters in the 2.5 km matrix are more mixed in color than in the spatial matrix, showing the combination of spatial proximity and semantic long-range connectivity.

It was found that incorporating the semantic matrix as additional information leads to improvements for most baselines, though DCRNN degrades significantly in certain configurations.
This work attributes this improvement to the long-range spatial connections formed by the semantic context (e.g., connecting functionally similar regions).
This concept is similar to that of an adaptive adjacency matrix, in which connections across large distances can be observed~\cite{shao_decoupled_2022, wu_graph_2019}.
Nevertheless, even for these approaches, this work shows that an additional semantic adjacency matrix improves the prediction skill, indicating that knowledge infusion guides the prediction process with new perspectives. 

\section{Conclusion}
This work introduces a general pipeline for the \textbf{\textit{Knowledge Infusion}} from a general purpose KG into spatio-temporal predictive tasks.
This work defines a parameterized subgraph extraction method related to a region of interest, and uses KGE to construct semantic adjacency matrices that connect the original spatial entities of the prediction task.
As a proof of concept, this paper shows how the proposed method improves the spatio-temporal prediction of six traffic prediction approaches. 
Experiments have shown that additional semantic context improves results for most baselines, with notable architecture-specific exceptions such as DCRNN.
Overall, it shows that traffic prediction approaches seem sensitive to the choice of the adjacency matrix, which is an opportunity to include external knowledge into these approaches.
Examining the structure of the employed adjacency matrices more closely suggests that the semantic adjacency matrix combines the spatial context's characteristics with long-range connectivity between sensors farther apart.
Future directions will include the cross perspective fusion, extracting semantical knowledge from data and inducing it with approaches like the presented to improve the underlying task. 

\section*{Acknowledgments}
This work has been funded by the Deutsche Forschungsgemeinschaft (DFG, German Research Foundation) PNr. 501836407 (NFDI4Objects), and funded by DFG under Germany's Excellence Strategy – EXC 2150 – 390870439.
This research was supported in part by high-performance computing resources at the Kiel University Computing Centre.

\bibliographystyle{IEEEtran}
\bibliography{sample-base}

\end{document}